\pdfoutput=1

\documentclass[11pt]{article}

\usepackage{EMNLP2023}

\usepackage{times}
\usepackage{latexsym}

\usepackage[T1]{fontenc}

\usepackage[utf8]{inputenc}

\usepackage{microtype}

\usepackage{inconsolata}
\usepackage[flushleft]{threeparttable}

\usepackage{tablefootnote}
\usepackage{graphicx} 
\usepackage{amsmath} 

\usepackage{booktabs}       
\usepackage{multicol,multirow,makecell}

\usepackage[normalem]{ulem}
\usepackage{amsmath,amssymb}

\usepackage{enumitem}
\usepackage{tabularx,booktabs}
\usepackage{soul}

\newcommand{\ee}{\boldsymbol{e}}
\newcommand{\beps}{\boldsymbol{\epsilon}}
\newcommand{\boldzero}{\mathbf{0}}

\newcommand{\normal}{\mathcal{N}}

\title{Improving End-to-End Speech Processing by Efficient Text Data Utilization with Latent Synthesis}

\author{
\textbf{Jianqiao Lu}$^1$\footnotemark[1]\thanks{~~Leading co-authors with equal contribution.}\hspace{4px}\thanks{~~Work done during an internship at Huawei.}~, 
\textbf{Wenyong Huang}$^{2*}$,\\
\textbf{Nianzu Zheng}$^2$\textbf{,} 
\textbf{Xingshan Zeng}$^2$\textbf{,} 
\textbf{Yu Ting Yeung}$^2$ \& \textbf{Xiao Chen}$^2$\\
{$^1$The University of Hong Kong \ \ \ $^2$Huawei Noah’s Ark Lab}\\
{\normalsize\texttt{jqlu@cs.hku.hk, wenyong.huang@huawei.com}}
}

\begin{document}
\maketitle
\begin{abstract}
Training a high performance end-to-end speech (E2E) processing model requires an enormous amount of labeled speech data, especially in the era of data-centric artificial intelligence. However, labeled speech data are usually scarcer and more expensive for collection, compared to textual data.
We propose Latent Synthesis (LaSyn), an efficient textual data utilization framework for E2E speech processing models.
We train a latent synthesizer to convert textual data into an intermediate latent representation of a pre-trained speech model. These pseudo acoustic representations of textual data augment acoustic data for model training.
We evaluate LaSyn on low-resource automatic speech recognition (ASR) and spoken language understanding (SLU) tasks. 
For ASR, LaSyn improves an E2E baseline trained on LibriSpeech train-clean-100, with relative word error rate reductions over 22.3\% on different test sets. 
For SLU, LaSyn improves our E2E baseline by absolute  4.1\% for intent classification accuracy and 3.8\% for slot filling SLU-F1 on SLURP, and absolute 4.49\% and 2.25\% for exact match (EM) and EM-Tree accuracies on STOP respectively.
With fewer parameters, the results of LaSyn are competitive to published state-of-the-art works. 
The results demonstrate the quality of the augmented training data. 
\end{abstract}

\section{Introduction}
In the data-centric artificial intelligence era, large quantity and high quality training data are essential for good performance of natural language processing (NLP) models including speech processing models.
A conventional speech processing system is usually cascaded with an automatic speech recognition (ASR) module and an NLP module. For example, in spoken language understanding (SLU) which predicts semantic information from speech input, the system first transcribes input speech into text with ASR, then pipes the text output to the natural language understanding (NLU) model for text analysis. 
An end-to-end (E2E) speech processing system leverages a single model which takes the input speech and performs spoken language processing tasks simultaneously. 
E2E models draw increasing attention due to less computational complexity and error propagation mitigation \cite{Shen2021ACL,Tian2020IESICR,Sharma2021ICASSP,LUGOSH2020ICASSP,WANG2020ICASSP,CHEN2021ICASSP}.
However, a challenge of E2E model training is the collection of enormous annotated spoken data, which are significantly more expensive to collect compared with the text-only counterpart. 
In contrast, for a cascaded system, the ASR module and NLP module are trained separately with paired speech-transcription data and annotated textual data respectively. Separated types of data are usually more readily available and thus lower data collection costs.
As the amount of high quality training data is critical for an E2E model,
a strategy to alleviate the inadequate spoken data problem with more abundant textual data. 

Two approaches have been proposed for utilizing textual data for E2E speech models in the literature. The first is modality conversion which utilizes a text-to-speech (TTS) system to convert text into speech \cite{laptev2020you}. The disadvantage is the requirement for a high-quality expressive TTS system. Another approach is unified representation learning for matching latent representations of speech and text with alignment losses \cite{bapna2021slam, Chen2022MAESTRO}. Given the significant difference between speech and text, aligning the hidden latent space of the two modalities is challenging.

We propose Latent Synthesis (LaSyn), a method to utilize text-only data for E2E speech processing models. LaSyn can be seen as an integration of the above two ideas. 
We train a latent synthesis model which synthesizes textual data into an intermediate latent representation of a pre-trained speech model. 
Compared to modality conversion, speech latent representation contains fewer details and redundancy than the original speech signal, thus is easier to synthesize.
Compared to unified representation learning, instead of aligning two modalities of huge difference, LaSyn learns to map the text into the latent representation of speech directly.

We evaluate LaSyn on low-resource ASR and SLU tasks. 
Low-resource ASR has gained big progress with the advancement of self-supervised speech pre-training \cite{Baevski2020ANIPS, Hsu2021ITASL, huang2022spiral}.
Further performance improvement still relies on external language models \cite{Baevski2020ANIPS}. 
We show that LaSyn allows an E2E ASR model to utilize text-only data effectively without external language models, and outperforms ASR models with external language models. We further evaluate LaSyn on two publicly available datasets for SLU tasks, namely SLURP  \cite{Bastianelli2020EMNLP} and Spoken Task Oriented Semantic Parsing (STOP) \cite{Tomasello2022STOP}. LaSyn achieves comparable performance to the state-of-the-art (SOTA) SLU models but with significantly fewer model parameters. 
We summarize our contributions as follows:
\begin{itemize}[leftmargin=*]

\item{We propose LaSyn, an efficient  textual  data  utilization  framework  for  E2E  speech  processing models. The framework enables cross-modal knowledge transfer from text to E2E speech processing models through latent synthesis. }
\item{We design 2 implementations for latent synthesizer which is the core of LaSyn framework: a fixed-projection latent synthesizer, and a diffusion latent synthesizer which applies recent progress of generative model, diffusion probabilistic model \cite{ho2020denoising, song2020score}.}

\item{
By improving an E2E ASR model through textual data utilization with LaSyn, we achieve competitive results on a low-resource ASR setup than published supervised ASR models which utilize textual data through an external language model.}

\item{
With LaSyn, we demonstrate E2E SLU models can be improved with a diverse set of textual NLP tasks, including NLU, information extraction (IE), named entity recognition (NER), and masked language modeling (MLM).
We achieve competitive results to published SOTA works on two publicly available SLU datasets, with significantly fewer model parameters. 
}

\end{itemize}

This paper is organized as follows. In the next section, we discuss related works of LaSyn. In Section \ref{method}, we discuss the model structure and training of LaSyn. We present experimental setup and results in Section \ref{expeirments}, and ablation studies on SLU tasks in Section \ref{ablation}. Finally, we conclude our work in Section \ref{conclusion}.

\section{Related Works}
\label{related}
In this section, we discuss the prior works of modality conversion and unified representation learning  related to LaSyn.

\begin{description}[style=unboxed,leftmargin=0cm]
\item[Modality conversion:] \citet{laptev2020you} shows that TTS data augmentation improves ASR performance in a low-resource setting. \citet{sun2020generating} further shows that the diversity and quality of the TTS system are important for ASR data augmentation.
\citet{Chen2022ICASSP} demonstrates similar representations derived from synthesized speech help downstream ASR tasks.
\citet{LUGOSH2020ICASSP} confirms the effectiveness of speech synthesis for E2E SLU models, either as a sole source of training data or as a form of data augmentation.
\citet{Thomas2021ICASSP} utilizes artificially synthesized speech to adapt a SLU model based on a recurrent neural network transducer.
\citet{Huang2020ICASSP} demonstrates the effectiveness of a multi-speaker TTS system under a low-resource SLU setting.
\citet{kharitonov2023speak} decouples the text-to-semantic and semantic-to-acoustic tasks to realize a multi-speaker text-to-speech system.
LaSyn generates pseudo acoustic representations from text without requiring a vocoder for speech waveform generation.

\item[Unified representation learning:]
\citet{Ao2021SPEECHT5} extends the idea of T5 \cite{Raffel2020JMLR} and proposes Speech-T5 with a cross-modal vector quantization in a shared discrete latent space.
\citet{Kim2021ICASSP} learns multi-modal alignment with two cross-modal pre-training tasks of masked language modeling and conditioned language modeling. 
\citet{Qian2021ICASSP}  unifies a pre-trained ASR encoder for speech and a pre-trained language model encoder for text into a transformer decoder.
\citet{sato2022text} introduces an adaptation branch to embed acoustic and linguistic information in the same latent space. 
\citet{thomas2022integrating} trains an RNN-T model both on speech and text inputs.
\citet{Zhang2022SPEECHLM}  introduces two alternative discrete phoneme-unit and hidden-unit tokenizers to bridge speech and text modalities.
MAESTRO \cite{Chen2022MAESTRO} learns unified 
representations of text and speech through sequence matching and duration prediction.
\citet{Chung2018NIPS} attempts to align the individually learned  text and speech  embedding via adversarial training and a refinement procedure.
SpeechUT \cite{Zhang2022EMNLP} leverages hidden units as the bridge between the speech encoder and the text decoder.
SpeechGPT\cite{zhang2023speechgpt} applies modality-adaptation pertaining and cross-modal instruction fine-tuning to perceive and generate multi-model content.
LaSyn connects text and speech information by mapping text representation directly into the pseudo acoustic latent space of a pre-trained speech model.
\end{description}

\begin{figure}[tb] 
\centering
\includegraphics[width=0.5\textwidth]{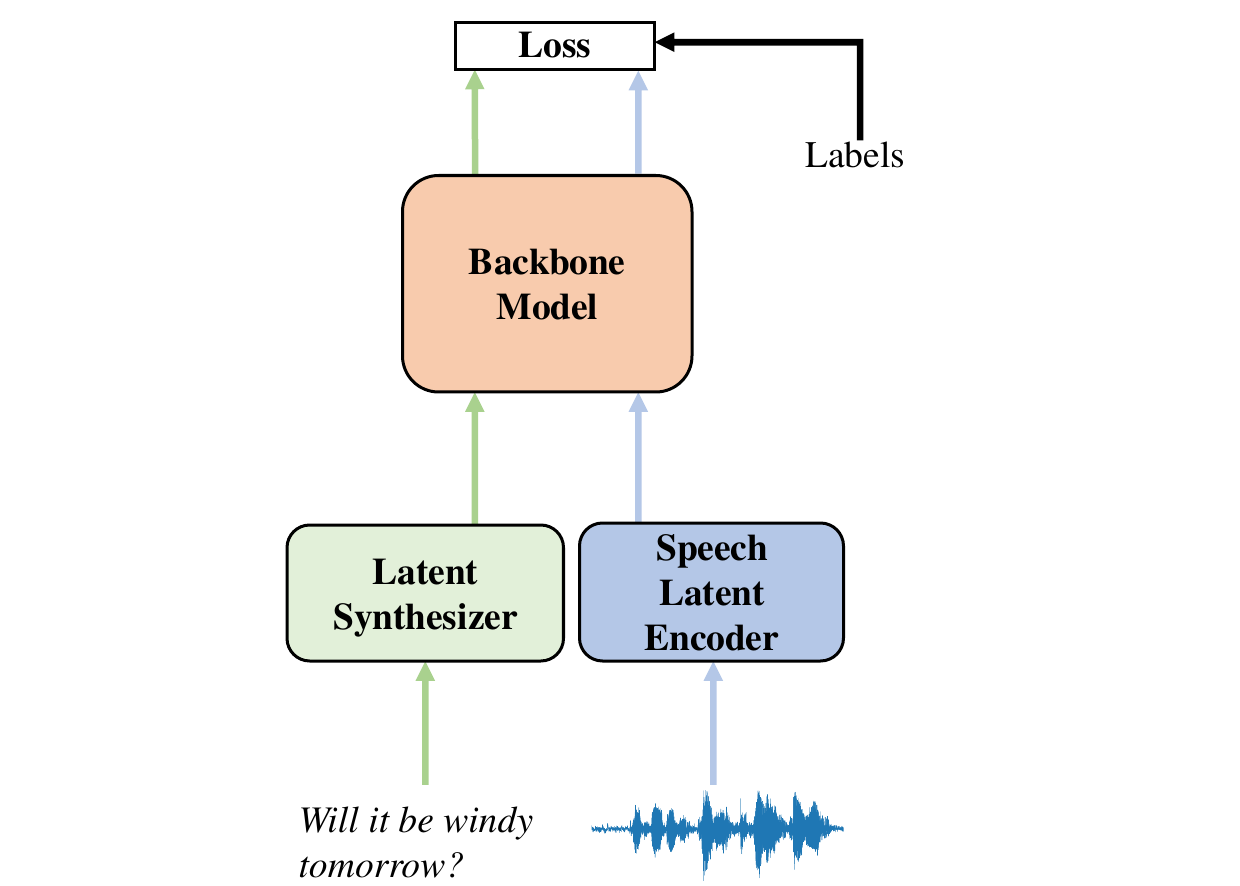}
\caption{The architecture of LaSyn framework.}
\label{Fig:overallarchitecture} 
\end{figure}

\section{Method}
\label{method}
\subsection{Architecture}
The LaSyn framework is illustrated in Fig. \ref{Fig:overallarchitecture}. The framework has 3 components: a speech latent encoder which maps speech data to corresponding speech latent representation, a latent synthesizer that projects text into the speech latent space, and a backbone model which is trained with either speech latent representations or pseudo acoustic latent representations from text.

\subsection{Training procedure}
\subsubsection{Speech Latent Encoder} 

Speech latent encoder is obtained from a pre-trained speech processing model, which is a supervised ASR model as illustrated in Fig. \ref{Fig:guiding_net} in this work. The parameters of speech latent encoder are frozen in the latter training stages to fix the speech latent space.

\subsubsection{Latent Synthesizer} 
We then train a latent synthesizer to project textual data into the same speech latent space of the speech latent encoder. 
Latent synthesizer allows utilizing training samples from textual data, which is
the core of the LaSyn framework. 
We explore two implementations of the latent synthesizer.

\begin{figure}[tb] 
\centering 
\includegraphics[width=0.42\textwidth]{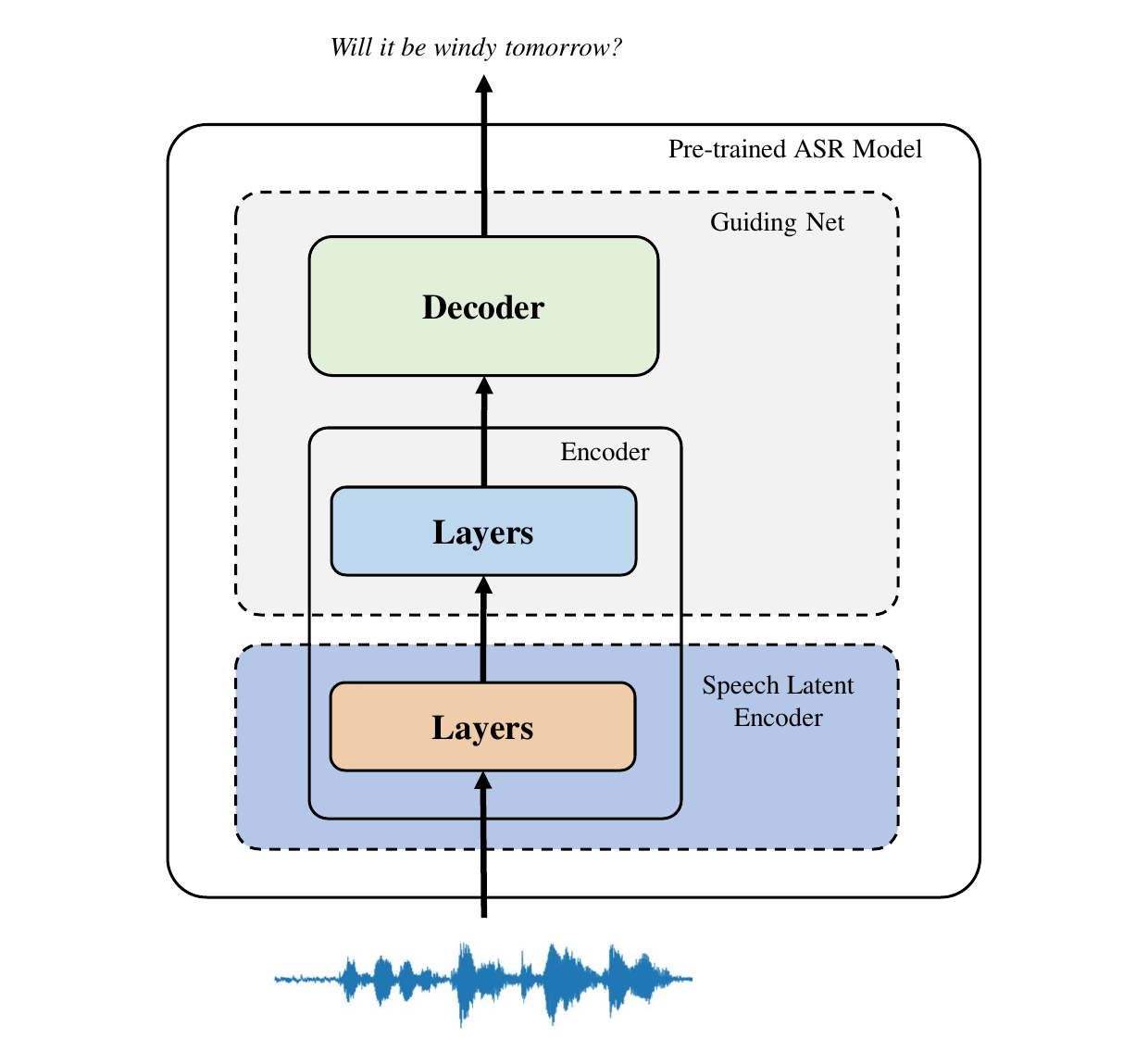}
\caption{Speech Latent Encoder and Guiding Net from a pre-trained ASR model.} 
\label{Fig:guiding_net}
\end{figure}

\begin{figure}[thb] 
\centering 
\includegraphics[trim={0 0 0 2.5cm},clip, width=0.42\textwidth]{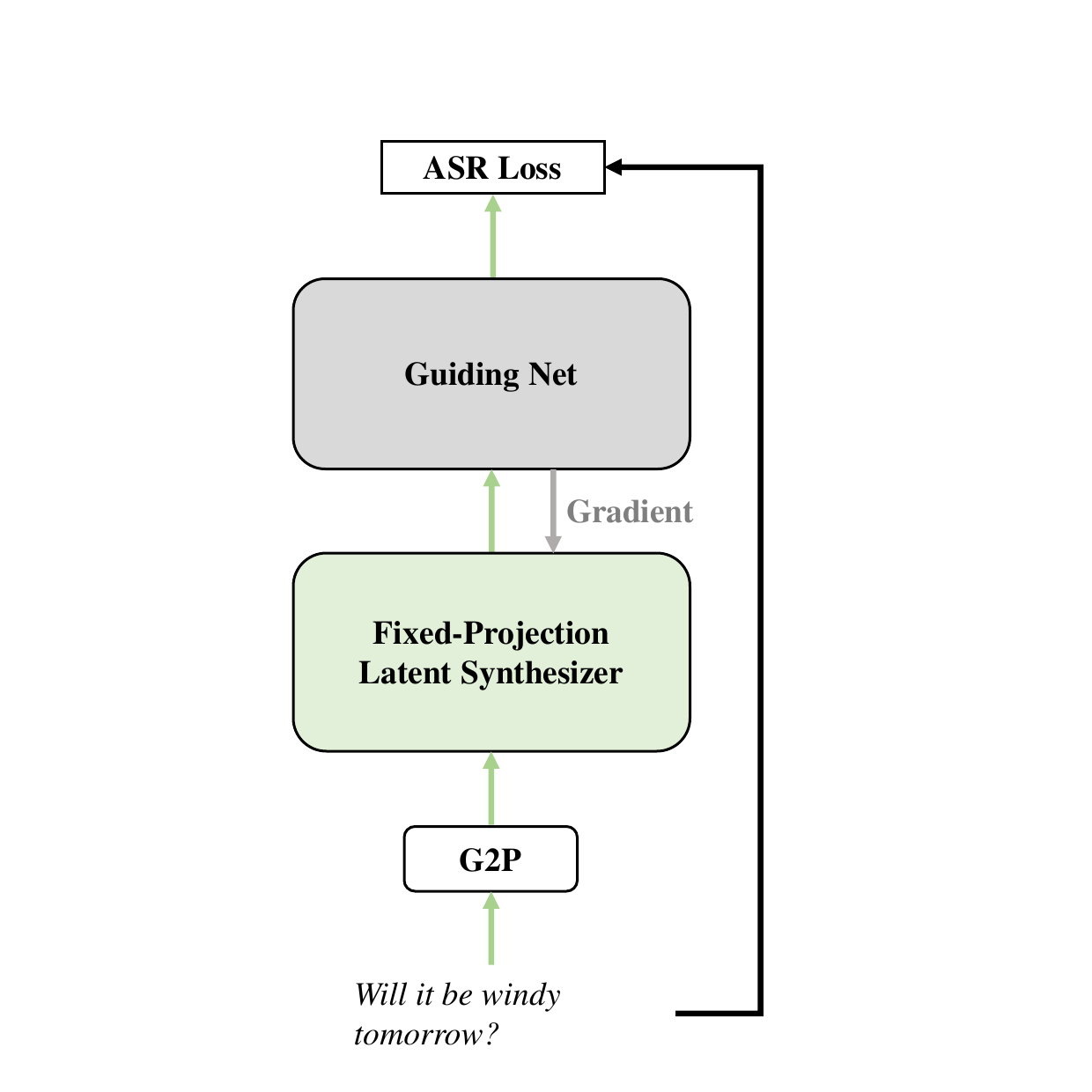}
\caption{Training process of Fixed-Projection Latent Synthesizer. We freeze parameters of Guiding Net.} 
\label{Fig:fixed_projection} 
\end{figure}

\begin{figure}[tb] 
\centering 
\includegraphics[width=0.495\textwidth]{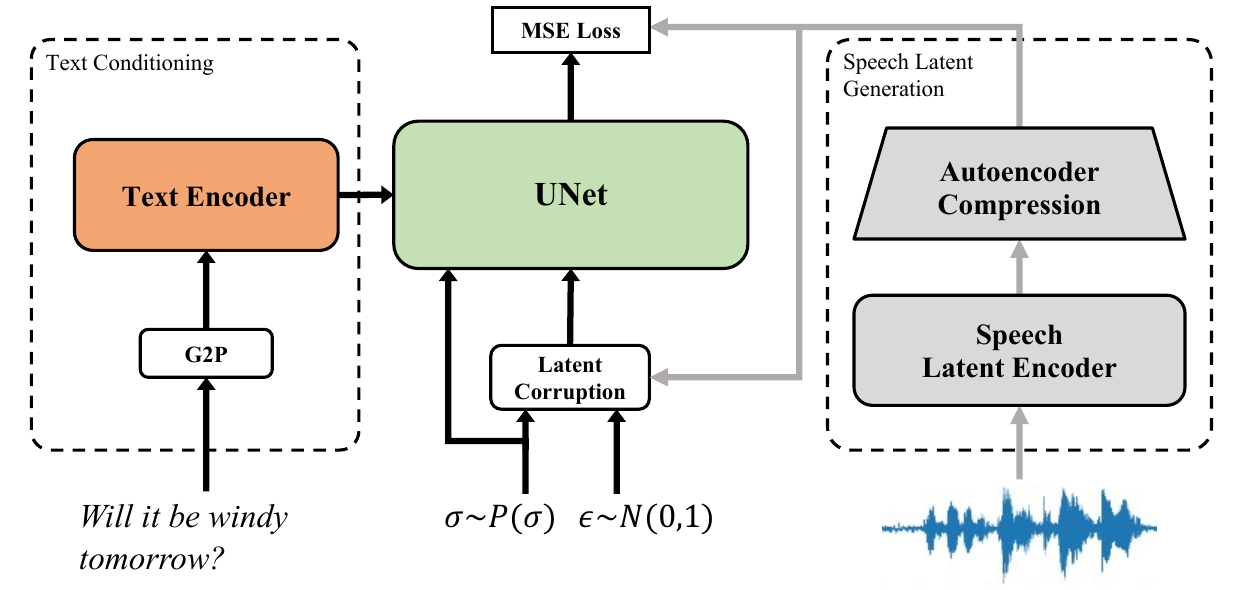}
\caption{Diffusion Latent Synthesizer training. The gray color indicates that the Speech Latent Encoder and Autoencoder are frozen during training.} 
\label{Fig:diffusion} 
\end{figure}

\begin{description}[style=unboxed,leftmargin=0cm]
   \item [Fixed-projection Latent Synthesizer:] We train a fixed-projection latent synthesizer with the help of a guiding net. The guiding net is also obtained from the pre-trained ASR model as illustrated in Fig. \ref{Fig:guiding_net}. Note that the guiding net is frozen in this stage. The training procedure is illustrated in Fig. \ref{Fig:fixed_projection}. 
    We optimize a fixed-projection latent synthesizer to generate latent representations which are recognizable as input of the guiding net. As the name suggests, the fixed-projection latent synthesizer learns a fixed one-to-one projection between text data and speech latent representation. The training objective is defined as follows,

\begin{equation}
\begin{aligned}
     &\underset{\phi}{\operatorname{argmin}}~
    \mathcal{L}_{ASR}\bigg(G_\theta \big(P_{\phi} \big(\text{G2P}(t)\big)\big),t\bigg)
    \end{aligned}
    \label{equ:fixedprojection_training_objective}
\end{equation}
where $G_\theta$ and $P_{\phi}$  represent the guiding network and the fixed-projection latent synthesizer respectively, $\phi$ represents the parameters of the latent synthesizer, $t$ is the text input, and $\text{G2P}$ is a grapheme-to-phoneme module. 
$\mathcal{L}_{ASR}$ is the same loss function of the pre-trained ASR model, such as transducer loss \cite{grave2012} or cross-entropy loss for attention-based encoder-decoder (AED) \cite{Vaswani2017}. 

\item [Diffusion Latent Synthesizer:] We also experiment with diffusion probabilistic models (DPM) \cite{ho2020denoising} as the latent synthesizer. DPMs have achieved great success in TTS \cite{popov2021grad, chen2021wavegrad} and text-conditioned image synthesis \cite{nichol2021glide, saharia2022photorealistic} recently. 
We use the formulation of DPM proposed in \citet{karras2022elucidating}.
Diffusion latent synthesizer generates latent representations by sampling an initial latent representation from a noise distribution and iteratively denoising the sample using a denoising model $D(h_{noisy}; \ee, \sigma)$ where $h_{noisy}$ represents the noisy latent at the current step, $\ee$ denotes the conditional text.
The denoising model is composed of an UNet \cite{ronneberger2015u} and a text encoder as shown in Fig. \ref{Fig:diffusion}. To reduce the complexity of the diffusion model, we train an autoencoder to compress the latent representation and use the lower-dimensional latent representation as the target of the diffusion latent synthesizer, similar to \citet{rombach2022high}. For succinctness, we do not depict the training of autoencoder in Fig. \ref{Fig:diffusion}.
   The training objective is to minimize,
\begin{equation}
    \mathbb{E}_{p(h,\ee),p(\beps), p(\sigma)}\bigg[\lambda(\sigma)\big\Vert D(h + \sigma \beps;\ee, \sigma) - h \big\Vert^2_2\bigg]
     \label{equ:diffusion_training_objective}
\end{equation}

\noindent where $h$ is clean latent representation, $p(h,\ee)$ represents the training data distribution of latent-text pairs. The latent-text pairs are derived from a paired speech-text dataset and a speech latent encoder which converts the speeches into latent representations.
$p(\sigma)$ is the distribution of noise levels that defines the corruption schedule \cite{karras2022elucidating}. $p(\beps) \in \normal(\boldzero, \bold1)$ is the standard normal distribution, $\lambda(\sigma)$ is the weighting factor of noise levels.
We employ classifier-free diffusion guidance \cite{ho2022classifier} to control latent quality and text alignment when sampling from the diffusion latent synthesizer.
\end{description}

\subsubsection{Backbone Model and Dual-modality Training}
After we train the latent synthesizer, we train the backbone model. 
We freeze the speech latent encoder and the latent synthesizer during backbone model training. We utilize both text and speech data in training. 
The backbone model takes input latent features from either speech latent encoder or latent synthesizer. We formulate both text-to-text and speech-to-text tasks as a unified sequence-to-sequence problem and refer to as dual-modality training.  The training loss is specific to each task, i.e., transducer loss for ASR, and cross-entropy loss for SLU.
The amount of textual data is usually significantly larger than speech data. We first train the backbone model with textual data.
Then we train the backbone model with both text and speech data. 

\section{Experiments}
\label{expeirments}
\subsection{Training Data}
\subsubsection{ASR}
We apply a 100-hour subset (train-clean-100) of LibriSpeech~\cite{panayotov2015librispeech} as low-resource labeled speech data. We use the transcription of the whole 960-hour LibriSpeech training split (LS-960) as text-only data.
\subsubsection{SLU}
We evaluate LaSyn on two challenging SLU datasets, SLURP \cite{Bastianelli2020EMNLP} and STOP \cite{Tomasello2022STOP}.
SLURP is substantially larger and linguistically more diverse than previous SLU datasets.
STOP is a recently released dataset that is the largest and the most complex SLU dataset.
We also leverage a diverse set of  NLP text datasets from different tasks, including natural language understanding (NLU), named entity recognition (NER), and information extraction (IE).
The extra NLP text datasets are listed in Table \ref{tab:textdatasets}.
\begin{table}[tb]
\centering
\scalebox{0.75}{
\begin{tabular}{ll}
 \midrule
\textbf{Task} & \textbf{Dataset} \\
 \midrule
 \midrule
    & \text{CLINC150} \small{\cite{Larson2019Arxiv}} \\
    & \text{Redwood} \small{\cite{Larson2022REDWOOD}} \\
    & \text{GOOGLE-DSTC8} \small{\cite{Rastogi2020DSTC8}} \\
   & \text{Leyzer} \small{\cite{Sowanski2020ICTSD}} \\
    & \text{HINT3} \small{\cite{Arora2020EMNLP}} \\
    
      NLU  & \text{Chatbot-Corpus} \small{\cite{BraunEtAl2017ACL}} \\
     
   & \text{MultiWOZ} \small{\cite{Zang2020ACL}} \\
   & \text{BANKING77} \small{\cite{Casanueva2020ACL}} \\
   & \text{FEWSHOTWOZ} \small{\cite{Peng2020SCGPT}} \\
    & \text{ATIS} \small{\cite{Tur2010SLTW}} \\
    & \text{Schema}   \small{\cite{Rastogi2019ARXIV}} \\
 \midrule
    &  \text{CrossNER} \small{\cite{Liu2020CROSSNER}} \\
    &  \text{WNUT17} \small{\cite{Derczynski2017NUGT} }\\
NER
    &  \text{CoNLL-2003} \small{\cite{Sang2003CONLL2003}} \\
    &  \text{CoNLL-2004} \small{\cite{Carreras2004CCNLL}} \\

 \midrule

  \multirow{2}{*}{IE} &  \text{OntoNotes} \small{\cite{Weischedel2013LDC}} \\
  
    &  \text{SCIERC} \small{\cite{Luan2018EMNLP}} \\
 \midrule
\end{tabular}
}
\caption{Extra NLP datasets for SLU experiments.}
\label{tab:textdatasets}
\end{table}

\label{asr_setup}

\begin{table}[tb]
    \centering

    \setlength{\extrarowheight}{2pt} 
    \scalebox{0.75}{
    \begin{tabularx}{0.5\textwidth}{c c}
        \toprule
        Channel multiplier & $[1, 1, 1, 1]$ \\
        Dropout & $0.1$ \\
        Number of channels & $256$ \\
        Number of residual blocks & $1$ \\
        Self attention resolutions & $[4, 2]$ \\
        \bottomrule
    \end{tabularx}
    }
     \caption{ \label{tab:unet} Hyper-parameters of UNet model}
   
\end{table}

\begin{figure}[tb] 
\centering
\includegraphics[ width=0.25\textwidth]{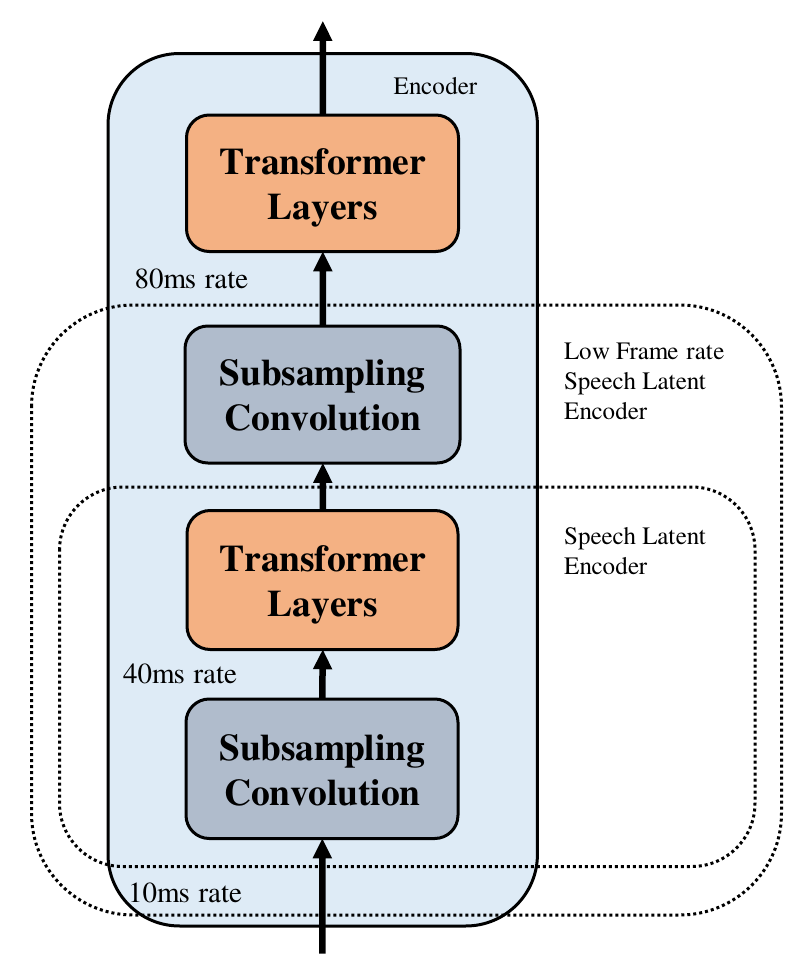}
\caption{Encoder architecture of the ASR model.  The frame rate of input is denoted as `10/40/80 ms'. } 
\label{Fig:lasyn_asr_encoder} 
\end{figure}

\begin{figure}[thb] 
\centering
\includegraphics[ width=0.3\textwidth]{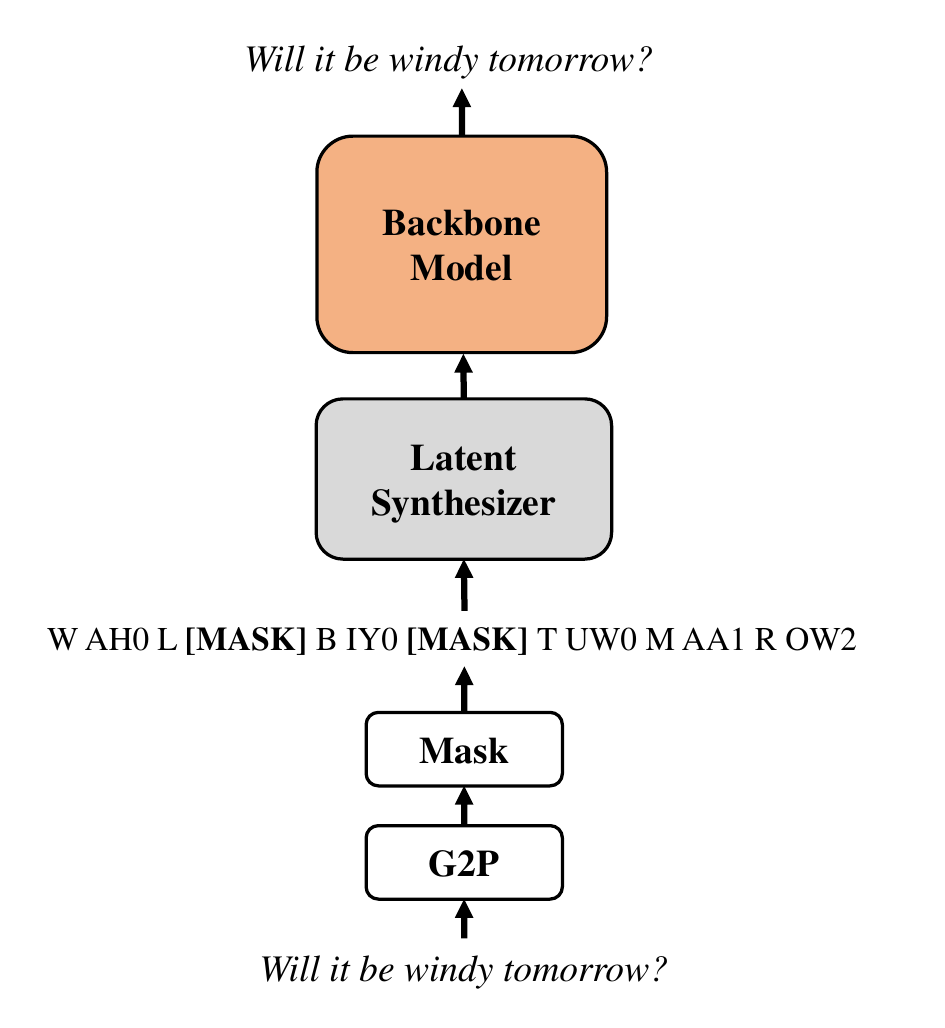}
\caption{\label{Fig:mlm_task} MLM task for utilizing unlabeled text data. [MASK] denotes the masked position.} 
\end{figure}

\subsection{Model and Training Setups}

\subsubsection{ASR}

For ASR pre-training, we use a Transformer Transducer model \cite{tian2019self,yeh2019transformer,zhang2020transformer}. We apply a 128-dimensional log-mel filterbank with 20~ms window length and 10~ms frame rate as input acoustic feature. We interleave strided-convolutions in the encoder to gradually down-sample the input speech as illustrated in  Fig. \ref{Fig:lasyn_asr_encoder}, which reduces computation effectively with negligible performance degradation \cite{Peddinti2018, han20_interspeech, huang2020interspeech}.
This model is pre-trained with train-clean-100.
SpecAugment \cite{park2019specaugment} is applied to avoid overfitting.
This pre-trained model is also our E2E ASR baseline. 
We obtain a speech latent encoder from this pre-trained model.

For latent synthesizers, we evaluate both fixed-projection latent synthesizer and diffusion latent synthesizer.
The fixed-projection latent synthesizer is composed of 4 1-D convolutional layers of 512 filters with a kernel size of 5. We observe that a simple model structure is sufficient. 
We train the diffusion latent synthesizer with train-clean-100. The text encoder is composed of two convolution layers followed by the two-layer transformer. The number of channels is 256.
The UNet model is adapted for 1-D sequence processing. The hyper-parameters of the UNet model are listed in Table \ref{tab:unet}. We use a small model such that
the latent synthesizer generates the pseudo acoustic latent representations on the fly during dual-modality training.

The backbone model is the same as the guiding net in Fig. \ref{Fig:guiding_net}. To utilize textual data in dual-modality training of the backbone model, we design a task similar to masked language modeling (MLM) \cite{devlin2018bert} as illustrated in Fig. \ref{Fig:mlm_task}. We randomly mask 30\% of input phonemes converted by g2pE\footnote{https://github.com/Kyubyong/g2p} according to CMUDict\footnote{https://github.com/cmusphinx/cmudict},  and train the backbone model to predict the corresponding words. 

We note that the parameters of the guiding net are frozen in latent synthesizer training. If we do not provide textual data for backbone model training, we just update the E2E baseline with extra epochs with a frozen speech latent encoder.

\subsubsection{SLU}
\begin{figure}[tb] 
\centering 
\includegraphics[width=.8\columnwidth]{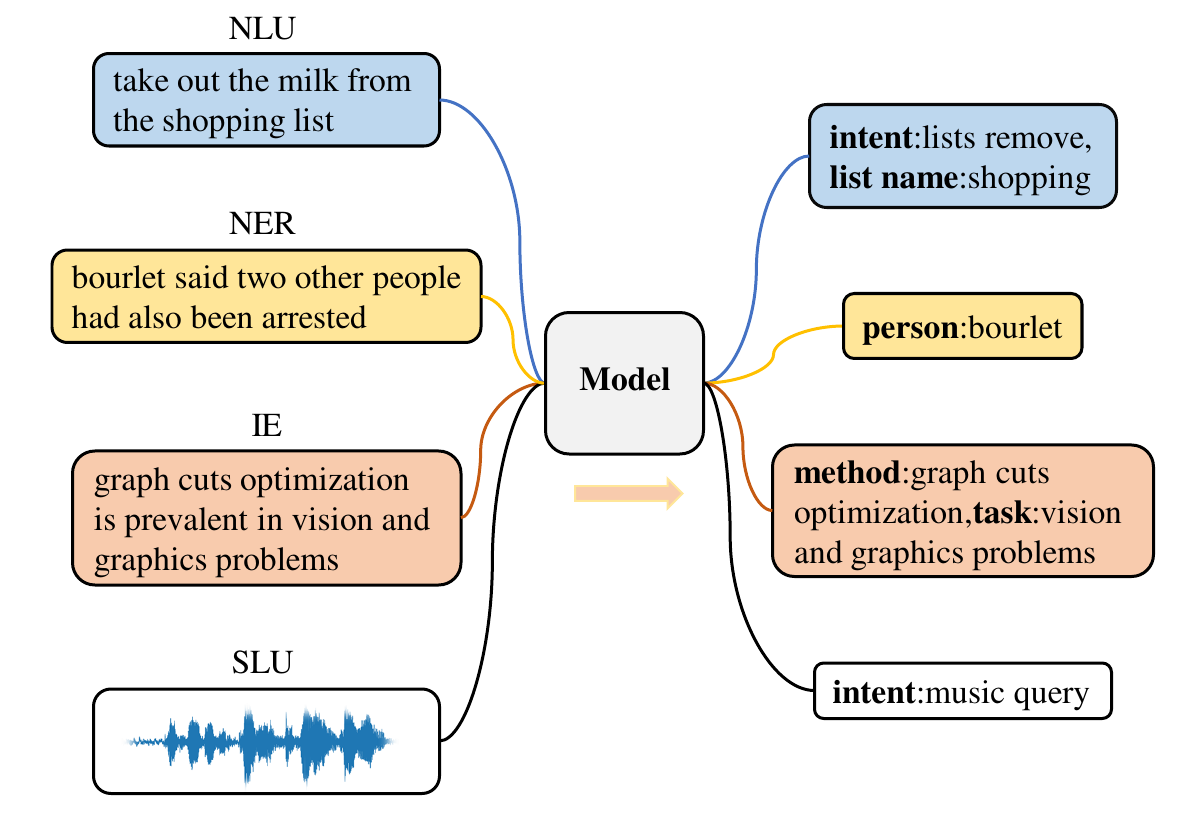}
\caption{Dual-modality training for SLU with LaSyn. The output labels of different tasks are converted to text sequences as shown in the right blocks. Meta values such as slot type and entry type are in bold.}
\label{Fig:multitasklearning} 
\end{figure}

We apply an attention-based encoder-decoder model for ASR pre-training. The  pre-trained ASR model is trained with LS-960 and SLURP speech data. The structure of the encoder is similar to the one in ASR experiments described in section \ref{asr_setup}.
We apply a 6-layer, 256-dimensional Transformer as the decoder. 
We evaluate the two implementations of the latent synthesizer similar to ASR experiments.
For fixed-projection latent synthesizer, the configuration is the same as ASR experiments. We apply text transcription of LS-960 for training.
For diffusion latent synthesizer, we use LS-960 as paired speech-text training data.
The backbone model shares the same model structure as the guiding net in Fig. \ref{Fig:guiding_net}. We also initialize the parameters from the guiding net. We train the backbone model with multiple tasks, including SLU, NLU, NER, and IE.
We convert the annotation of all the datasets to a text-sequence format as illustrated in Fig. \ref{Fig:multitasklearning}. We formulate all the tasks as a unified sequence-to-sequence problem.

We note that the model structure of the E2E baseline model is the same as the LaSyn model, but the latent synthesizer is disabled. The E2E baseline model does not train with any additional textual data.
We fine-tune the E2E baseline model with SLU task after ASR pre-training.

\begin{table}[tb]
\centering
{ \scalebox{0.65}{
 \begin{tabular}{lcrr}
 \toprule
 \multirow{2}{*}{\textbf{Model}} & \multirow{2}{*}{\textbf{LM}} & \multicolumn{2}{c}{\textbf{test}} \\
 \cline{3-4}
 {} & {} & \textbf{clean} & \textbf{other} \\
 \midrule
 \midrule
 Hybrid DNN/HMM \small{\cite{L_scher_2019}} & 4-gram & \large{5.8} & \large{18.6} \\

 LAS~\small{\cite{park2020improved}} & LSTM & \large{5.5} & \large{16.9} \\
 
 Conformer-CTC \small{\cite{Watanabe2018EspnetLibrispeech100h}} & - & \large{7.7} & \large{20.6} \\
 Conformer-CTC/Attention \small{\cite{Watanabe2018EspnetLibrispeech100h}} & - & \large{7.3} & \large{19.3} \\
 Conformer-Transducer \small{\cite{Watanabe2018EspnetLibrispeech100h}}  & - & \large{7.8} & \large{19.8} \\
\midrule\midrule
 TTS data augm.~\small{\cite{laptev2020you}} & - & \large{6.8} & \large{19.9} \\
 TTS data augm.~\small{\cite{laptev2020you}} & LSTM & \large{\textbf{4.3}} & \large{\textbf{13.5}} \\
\midrule\midrule
 E2E baseline (ours)  & - & \large{7.4} & \large{20.1} \\
 LaSyn-FixedProj-LFR (ours)   & - &    \large{4.5} & \large{17.1} \\
 LaSyn-FixedProj (ours)   & - &    \large{4.5} & \large{16.1} \\
 LaSyn-Diffusion (ours)  & - &    \large{\textbf{4.4}} & \large{\textbf{15.6}} \\

 \bottomrule
 \end{tabular}
 }
 }
  \caption{Low-resource ASR results trained with train-clean-100 split of LirbiSpeech.
 We compare LaSyn with published supervised methods. 
     We report WER (\%) on dev/test sets.}
 \label{tab:low_resource_asr}
\end{table}

\subsection{ASR Results}
The experimental results of ASR are shown in Table \ref{tab:low_resource_asr}. 
We first compare LaSyn models with our E2E baseline which achieves comparable performance to conformer-based models. 
The only difference is that the LaSyn models are trained with additional textual data. The LaSyn-Diffusion model, which uses a diffusion latent synthesizer, achieves 40.5\% and 22.3\% relative WER reductions on test-clean and test-other of Librispeech test sets compared to the E2E baseline. We notice that the improvement on test-clean is more significant than test-other. 
Both the fixed-projection latent synthesizer and the diffusion latent synthesizer are trained with train-clean-100 which contains only clean speech.
We speculate that the limited variety of training data  train-clean-100 biases ASR performance toward clean speech.

We also observe that the performance of the model with fixed-projection latent synthesizer (LaSyn-FixedProj) is only slightly worse than LaSyn-Diffusion. The result is surprising, as the fixed-projection latent synthesizer is simpler than the diffusion latent synthesizer. The diffusion latent synthesizer may need further hyper-parameter tuning, or may need more training data for better performance. 
The LaSyn-FixedProj-LFR model utilizes a low frame rate speech latent encoder as illustrated in Fig. \ref{Fig:lasyn_asr_encoder}. The performance is slightly worse than the LaSyn-FixedProj on test-other.

Compared to published supervised ASR models that utilize text data through external language models, LaSyn models perform better without an external language model (LM).
Compared to the published methods using TTS for data augmentation, the performance of LaSyn models are significantly better without an external LM.
Given the existence of real-world scenarios with limited labeled speech data, such as minority languages and specific domains, our proposed method offers a novel approach to developing ASR applications.

\subsection{SLU Results}
\subsubsection{SLURP}
The experimental results of SLURP are shown in Table \ref{tab:thetotalresultonslurp}. We report accuracy for intent classification (IC), and SLU-F1\cite{Bastianelli2020EMNLP} for slot filling (SF). 

We first compare LaSyn models with our E2E baseline. 
Compared to the E2E baseline, LaSyn-FixedProj improves IC accuracy and SF SLU-F1 by  absolute 4.1\% and 3.8\% respectively. The result suggests that knowledge of textual NLP data is effectively transferred to SLU model.
LaSyn-Diffusion performs slightly worse than LaSyn-FixedProj. We believe that with further hyper-parameter tuning and more training data, the performance of diffusion latent synthesizer should be further improved.

We further compare the LaSyn models with previously published E2E SLU results. The published models are fine-tuned from HuBERT \cite{Hsu2021ITASL} Base (95~M parameters) or Large (300~M parameters).  The performance of LaSyn-FixedProj is comparable to ESPnet-SLU \cite{Arora2022ICASSP} and PF-hbt-base \cite{Wang2021HUBERTPARTIAL}.
The IC accuracy of LaSyn-FixedProj is slightly worse than EF-hbt-large \cite{Wang2021HUBERTPARTIAL}, but the number of parameters is 8 times fewer.

\begin{table}[tb]
 \centering
{
 \scalebox{0.7}{
 \begin{tabular}{llcc}
 \toprule

 \multirow{2}{*}{\textbf{Model}}  &   \multirow{2}{*}{\textbf{\# Params}}  &  {\textbf{IC}}  &  {\textbf{SF}}   \\
 {} & {} & \textbf{(ACC \%)} &  \textbf{(SLU-F1)} \\
 \midrule
 \midrule
 
 ESPnet-SLU  \small{\cite{Arora2022ICASSP}}   & $\geq$ 300 M  & 86.3 &71.9   \\
 PF-hbt-base \small{\cite{Wang2021HUBERTPARTIAL}}  &   $\geq$ 90~M  &  87.5 & 75.3    \\
 EF-hbt-large \small{\cite{Wang2021HUBERTPARTIAL}} & $\geq$ 300~M &  \textbf{89.4} & 78.4 \\
\midrule\midrule
E2E Baseline (ours)   & 37.8 M  & 84.4 &74.7 \\
LaSyn-Diffusion (ours) & 37.8~M &  87.4 &  77.3  \\
LaSyn-FixedProj (ours) & 37.8~M  & 88.5 & \textbf{78.5}     \\
 \bottomrule
 \end{tabular}
  }
 }
  \caption{ \label{tab:thetotalresultonslurp} Results on SLURP dataset. We report accuracy (ACC\%) for the IC task and SLU-F1 for the SF task. }

\end{table}

\begin{figure}[tb] 
\centering 
\includegraphics[width=0.45\textwidth]{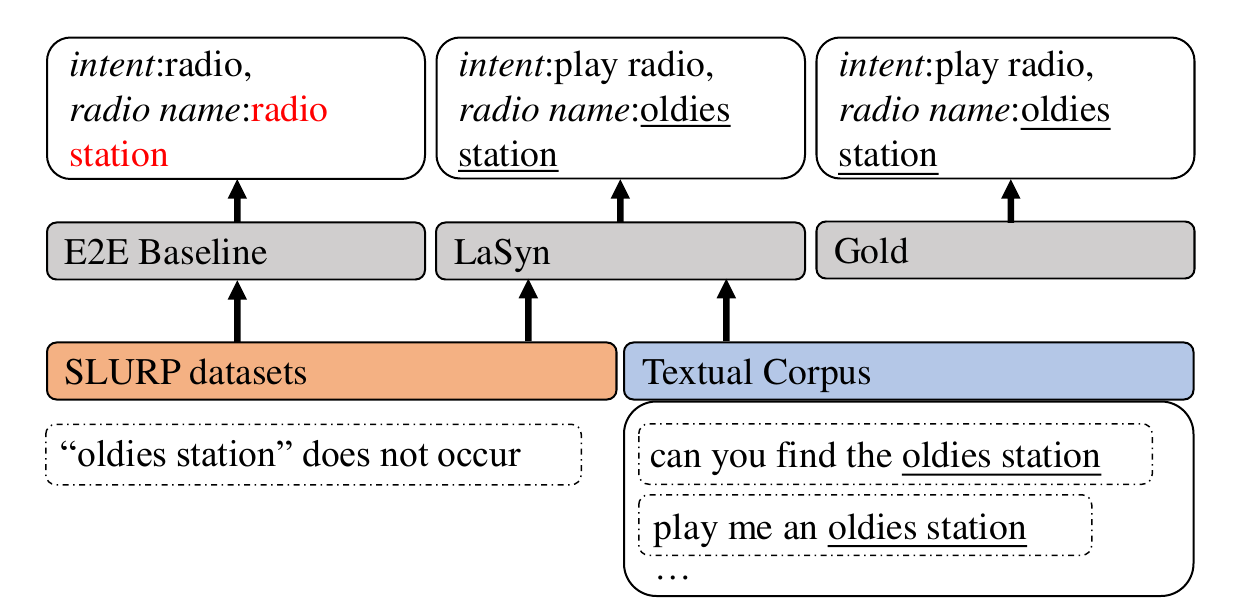}
\caption{An example of LaSyn output from SLURP test set. The target "oldies station" does not appear in SLU training data while LaSyn utilizes knowledge from the textual corpus. Meta values such as "intent" and slot type are \textit{italicized}.} 
\label{Fig:caseanalysis} 
\end{figure}

To understand how LaSyn improves our baseline E2E SLU model, we further analyze samples from the test set that LaSyn performs better than our baseline. An example is shown in Fig. \ref{Fig:caseanalysis}.
Our E2E baseline model fails for the slot "Oldies Station", as this phrase never occurs in the SLURP training set.
In contrast, LaSyn model correctly predicts the slot value. This phrase is included in the textual corpora. The text knowledge is transferred to SLU model with the LaSyn framework.
The baseline E2E SLU model does not get the proprietary term `Oldies Station' while LaSyn predicts this unique vocabulary successfully.

\subsubsection{STOP}

We present our results of STOP in Table \ref{tab:thetotalresultonstop}. Compared to our E2E baseline, LaSyn-FixedProj improves EM accuracy and EM-Tree accuracy on the test set by absolute  4.49\% and 2.25\% respectively, again suggesting that there is effective cross-modality text knowledge transfer. 

We further compare our results with STOP-E2E and STOP-Cascaded \cite{Tomasello2022STOP}. STOP-E2E is an encoder-decoder based Transformer model fine-tuned from an E2E ASR model. The E2E ASR model is fine-tuned from HuBERT Base \cite{Hsu2021ITASL}. STOP-Cascaded is a cascaded system composed of an ASR system fine-tuned from wav2vec2.0 Base \cite{Baevski2020ANIPS} and an NLU model fine-tuned from a BART Base model \cite{lewis2019bart}.
LaSyn-FixedProj performs slightly better than STOP-E2E with 0.25\% and 2.63\% absolute improvement of EM and EM-Tree accuracies on the test set respectively.
However, compared to STOP-Cascaded on the test set, while LaSyn-FixedProj is competitive on EM-Tree accuracy, EM accuracy is slightly inferior. The number of parameters in LaSyn models is much fewer. We expect performance improvement with more model parameters. 

\begin{table}[tb]
 \centering
{
\scalebox{0.62}{
 \begin{tabular}{llcc}
 \toprule
 
\multirow{2}{*}{\textbf{Model}} & \multirow{2}{*}{\textbf{\# Params}}&    \textbf{dev} & \textbf{test}  \\

 {} &  {} & \textbf{EM / EM-Tree} & \textbf{EM / EM-Tree}  \\

 \midrule
 \midrule

STOP-E2E    & \multirow{2}{*}{$\geq$ 90~M} &  \multirow{2}{*}{69.12 / 83.89}  &  \multirow{2}{*}{69.23 / 82.87} \\
\small{\cite{Tomasello2022STOP}} & & & \\
STOP-Cascaded   &  \multirow{2}{*}{$\geq$ 230~M} & \multirow{2}{*}{72.43 / 86.58}  &  \multirow{2}{*}{\textbf{72.36} / \textbf{85.77}}  \\
\small{\cite{Tomasello2022STOP}} & & & \\
\midrule\midrule
  E2E Baseline (ours) &  37.8~M &   64.02 / 82.84  &  64.99 / 82.25 \\
  LaSyn-Diffusion (ours) & 37.8~M & 67.91 /  85.57 &  68.33 / 84.92     \\
  LaSyn-FixedProj (ours) & 37.8~M & 69.33 /  86.24  &   \textbf{69.48} / \textbf{85.50}      \\
 \bottomrule
 \end{tabular}
 }
 }
  \caption{  \label{tab:thetotalresultonstop}Results on STOP dataset. We report the EM and EM-Tree accuracies (\%) on dev and test sets.}

\end{table}

\section{Ablation Study}
\label{ablation}
\subsection{Training with Unlabeled Textual Data}
Plain text data without annotation are more abundant than annotated NLP data. We experiment with SLU training with unlabelled textual data. We prepare the unlabelled text data by striping the annotation labels of the NLP datasets and keeping the input text. We apply the MLM task described in section \ref{asr_setup} to utilize the unlabeled textual data. We evaluate LaSyn models with fixed-projection latent synthesizer.
The results are listed in Table \ref{tab:ablation_annotation}.

The results show that LaSyn still benefits from unlabeled text, compared to our E2E baseline on both SLURP and STOP datasets. With unlabeled text and MLM tasks, LaSyn achieves an absolute improvement of 1.6 \% and 0.9 \% on IC and SF tasks on SLURP dataset, 2.19 \%, and 0.46\% on EM and EM-Tree on STOP test set.
While the improvement is not as significant as using labeled textual data, data collection is further simplified with unlabelled textual data.
\begin{table}[tb]
  \centering
{ \scalebox{0.56}{
 \begin{tabular}{llccc}
 \toprule
 
 \multirow{2}{*}{\textbf{Model}}  & \multirow{2}{*}{\textbf{Text Data}}  & \textbf{ SLURP}  & \textbf{STOP (dev)}  & \textbf{STOP (test)}  \\
 & & \textbf{(IC / SF)} & \textbf{(EM / EM-Tree)} & \textbf{(EM / EM-Tree)}\\
 \midrule
 \midrule
E2E Baseline  & - & 84.4 / 74.7  & 64.02 / 82.84 &  64.99 / 82.25 \\
LaSyn-FixedProj & labelled & 88.5 / 78.5 & 69.33 /  86.24  &   69.48 / 85.50       \\
LaSyn-FixedProj & unlabelled & 86.1 / 75.4 & 66.13 / 82.89 & 66.40 / 82.33     \\

 \bottomrule
 \end{tabular}
 }
 }
  \caption{Ablation study of unlabeled text data. We report results on SLURP test set, STOP dev and test sets. 
  }
 \label{tab:ablation_annotation}
\end{table}

\subsection{Training with Diverse NLP Tasks}
We do an ablation to observe the effect of training LaSyn with textual data from a diverse set of NLP tasks.
The results are shown in Table \ref{tab:stop_multi_task}. We observe that including each NLP task brings substantial improvement over the E2E baseline. As the NLU task is the most relevant to SLU, performance improvement is the most significant.
When we combine all the NLP tasks, there is marginal further performance improvement.

\begin{table}[tb]
  \centering
 \scalebox{0.6}{
 \begin{tabular}{llcc}
 \toprule
 
\multirow{2}{*}{\textbf{Model}} & \multirow{2}{*}{\textbf{Text Training data}}  &  \textbf{STOP (dev)} & \textbf{STOP (test)}  \\
& & \textbf{(EM / EM-Tree)} & \textbf{(EM / EM-Tree)}\\
 \midrule
 \midrule
{E2E baseline} & -  &  64.02 / 82.84 &  64.99 / 82.25 \\
 \midrule
\multirow{4}{*}{{LaSyn}} 
& {NLU}  &  68.99 / 86.31   &  69.40 / 85.45 \\
& {NER}  &  68.55 / 85.65   &  69.24 / 85.05 \\
& {IE}  &  68.43 / 85.50  &  68.88 / 84.99 \\
& {NLU + NER + IE} & 69.33 /  86.24  &   69.48 / 85.50     \\
 \bottomrule
 \end{tabular}
 }
  \caption{Results of LaSyn trained with text data of different NLP tasks. We report EM and EM-Tree accuracies (\%) on STOP dev and test sets.}
\label{tab:stop_multi_task}
\end{table}
\subsection{Latent Synthesizer as Acoustic Augmentation}
We experiment with using the fixed-projection latent synthesizer for acoustic augmentation. 
We extract the transcription and the annotation from the SLU dataset to form an NLU dataset. When training the backbone model, we apply both the SLU and the NLU datasets in dual-modality training.
As the NLU dataset is derived from the SLU dataset, the latent synthesizer does not introduce extra textual content. 
Pseudo speech latent representations from the latent synthesizer are considered as an augmentation of the original speech latent representation.

As shown in Table \ref{tab:lasynsluonly_slurpstop}, SLU performance improves significantly over the E2E baseline but does not reach the level of Table \ref{tab:stop_multi_task} which utilizes extra NLP datasets.     
Further enriching the diversity of pseudo acoustic latent is the potential to improve SLU performance.
\begin{table}[tb]
 \centering
 \scalebox{0.59}{
 \begin{tabular}{lccc}
 \toprule
 \multirow{2}{*}{\textbf{Model}}  &  \textbf{SLURP}& \textbf{STOP (dev)}  & \textbf{STOP (test)}  \\
 & \textbf{(IC / SF)}& \textbf{(EM / EM-Tree) } & \textbf{(EM / EM-Tree) } \\
 \midrule
 \midrule

E2E Baseline   &  84.4 / 74.7 & 64.02 / 82.84  &  64.99 / 82.25 \\

 LaSyn (Acoustic Aug.)   & 86.9 / 76.0 &  67.69 / 85.18 &  68.25 / 84.50     \\

 \bottomrule
 \end{tabular}
 }
  \caption{Results of acoustic augmentation with latent synthesizer. We report  IC (ACC\%) and SF (SLU-F1) for SLURP, EM and EM-Tree accuracies (\%) for STOP.}
\label{tab:lasynsluonly_slurpstop}
\end{table}

\section{Conclusion}
\label{conclusion}
We present LaSyn, a framework which enables efficient textual data utilization for E2E speech processing.
By converting text into pseudo acoustic latent representation with a latent synthesizer, cross-modality knowledge transfer from textual data to E2E speech processing models is achieved.
For the low-resource ASR task with Librispeech, LaSyn achieves relative WER reduction from 22.3\%  to  40.5\% on test sets, compared to our E2E baseline with the same model structure. The results are competitive to published works which utilize textual data through external language models.
For SLU tasks, LaSyn improves over our E2E baseline by absolute 4.1\% and 3.8\% for for IC accuracy and SF SLU-F1 on SLURP, and absolute 4.49\% and 2.25\% of EM and EM-Tree accuracies on STOP. The results are competitive to published SOTA works with much fewer model parameters.
Future improvement of latent synthesizer should further bridge the gap between speech and textual modality, which we leave as next step.

\section*{Limitations}
The core of our method is the generation of pseudo acoustic representation from text input. We focus on generating consistent latent sequences effectively.
We only evaluate two latent synthesis methods, including fixed-projection and diffusion latent synthesizers. There are other probable methods for latent generation, such as generative adversarial  networks (GAN) \cite{Goodfellow2020ACM}. 
Compared with TTS which generates audible speech suitable for human judgment, there is no subjective method to evaluate the quality and intelligibility of generated pseudo acoustic representation from the proposed framework, which is a main limitation. 
The design of reasonable quality indicators of acoustic representation would be meaningful for future work.
Moreover, we have not evaluated the proposed latent synthesis framework on other phonological systems such as tonal languages like Chinese. The effectiveness of the framework on tonal languages is not guaranteed.

\section*{Ethics Statement}
In this paper, we only use publicly available datasets for experiments.  Our experiments do not involve any subjective tests or human data annotations. In the experiments, the latent synthesis framework does not produce any audible speech content. We do not apply any specific speaker information during training and inference.

\bibliography{anthology,cite}

\begin{thebibliography}{74}
\expandafter\ifx\csname natexlab\endcsname\relax\def\natexlab#1{#1}\fi

\bibitem[{Ao et~al.(2021)Ao, Wang, Zhou, Liu, Ren, Wu, Ko, Li, Zhang, Wei et~al.}]{Ao2021SPEECHT5}
Junyi Ao, Rui Wang, Long Zhou, Shujie Liu, Shuo Ren, Yu~Wu, Tom Ko, Qing Li, Yu~Zhang, Zhihua Wei, et~al. 2021.
\newblock \href {https://arxiv.org/abs/2110.07205} {{SpeechT5}: Unified-modal encoder-decoder pre-training for spoken language processing}.
\newblock {arXiv preprint arXiv:2110.07205}.

\bibitem[{Arora et~al.(2020)Arora, Jain, Chaturvedi, and Modi}]{Arora2020EMNLP}
Gaurav Arora, Chirag Jain, Manas Chaturvedi, and Krupal Modi. 2020.
\newblock \href {https://www.aclweb.org/anthology/2020.insights-1.16} {{HINT}3: Raising the bar for intent detection in the wild}.
\newblock In \emph{Proceedings of the First Workshop on Insights from Negative Results in NLP}, pages 100--105.

\bibitem[{Arora et~al.(2022)Arora, Dalmia, Denisov, Chang, Ueda, Peng, Zhang, Kumar, Ganesan, Yan et~al.}]{Arora2022ICASSP}
Siddhant Arora, Siddharth Dalmia, Pavel Denisov, Xuankai Chang, Yushi Ueda, Yifan Peng, Yuekai Zhang, Sujay Kumar, Karthik Ganesan, Brian Yan, et~al. 2022.
\newblock {ESPnet-SLU}: Advancing spoken language understanding through {ESPnet}.
\newblock In \emph{2022 IEEE International Conference on Acoustics, Speech and Signal Processing (ICASSP 2022)}, pages 7167--7171.

\bibitem[{Baevski et~al.(2020)Baevski, Zhou, Mohamed, and Auli}]{Baevski2020ANIPS}
Alexei Baevski, Yuhao Zhou, Abdelrahman Mohamed, and Michael Auli. 2020.
\newblock wav2vec 2.0: A framework for self-supervised learning of speech representations.
\newblock In \emph{Advances in Neural Information Processing Systems}, volume~33, pages 12449--12460.

\bibitem[{Bapna et~al.(2021)Bapna, Chung, Wu, Gulati, Jia, Clark, Johnson, Riesa, Conneau, and Zhang}]{bapna2021slam}
Ankur Bapna, Yu-an Chung, Nan Wu, Anmol Gulati, Ye~Jia, Jonathan~H Clark, Melvin Johnson, Jason Riesa, Alexis Conneau, and Yu~Zhang. 2021.
\newblock \href {https://arxiv.org/abs/2110.10329} {{SLAM}: A unified encoder for speech and language modeling via speech-text joint pre-training}.
\newblock {arXiv preprint arXiv:2110.10329}.

\bibitem[{Bastianelli et~al.(2020)Bastianelli, Vanzo, Swietojanski, and Rieser}]{Bastianelli2020EMNLP}
Emanuele Bastianelli, Andrea Vanzo, Pawel Swietojanski, and Verena Rieser. 2020.
\newblock Slurp: A spoken language understanding resource package.
\newblock In \emph{Proceedings of the 2020 Conference on Empirical Methods in Natural Language Processing (EMNLP)}, pages 7252--7262.

\bibitem[{Braun et~al.(2017)Braun, Hernandez-Mendez, Matthes, and Langen}]{BraunEtAl2017ACL}
Daniel Braun, Adrian Hernandez-Mendez, Florian Matthes, and Manfred Langen. 2017.
\newblock \href {http://www.aclweb.org/anthology/W17-3622} {Evaluating natural language understanding services for conversational question answering systems}.
\newblock In \emph{Proceedings of the 18th Annual SIGdial Meeting on Discourse and Dialogue}, pages 174--185.

\bibitem[{Carreras and M{\`a}rquez(2004)}]{Carreras2004CCNLL}
Xavier Carreras and Llu{\'\i}s M{\`a}rquez. 2004.
\newblock \href {https://aclanthology.org/W04-2412} {Introduction to the {C}o{NLL}-2004 shared task: Semantic role labeling}.
\newblock In \emph{Proceedings of the Eighth Conference on Computational Natural Language Learning ({C}o{NLL}-2004) at {HLT}-{NAACL} 2004}, pages 89--97.

\bibitem[{Casanueva et~al.(2020)Casanueva, Temcinas, Gerz, Henderson, and Vulic}]{Casanueva2020ACL}
I{\~{n}}igo Casanueva, Tadas Temcinas, Daniela Gerz, Matthew Henderson, and Ivan Vulic. 2020.
\newblock \href {https://arxiv.org/abs/2003.04807} {Efficient intent detection with dual sentence encoders}.
\newblock In \emph{Proceedings of the 2nd Workshop on NLP for ConvAI - ACL 2020}.

\bibitem[{Chen et~al.(2021{\natexlab{a}})Chen, Zhang, Zen, Weiss, Norouzi, Dehak, and Chan}]{chen2021wavegrad}
Nanxin Chen, Yu~Zhang, Heiga Zen, Ron~J Weiss, Mohammad Norouzi, Najim Dehak, and William Chan. 2021{\natexlab{a}}.
\newblock \href {https://arxiv.org/abs/2106.09660} {{WaveGrad 2}: Iterative refinement for text-to-speech synthesis}.
\newblock {arXiv preprint arXiv:2106.09660}.

\bibitem[{Chen et~al.(2021{\natexlab{b}})Chen, Lu, Mottini, Li, Droppo, Du, and Zeng}]{CHEN2021ICASSP}
Yixin Chen, Weiyi Lu, Alejandro Mottini, Li~Erran Li, Jasha Droppo, Zheng Du, and Belinda Zeng. 2021{\natexlab{b}}.
\newblock Top-down attention in end-to-end spoken language understanding.
\newblock In \emph{2021 IEEE International Conference on Acoustics, Speech and Signal Processing (ICASSP 2021)}, pages 6199--6203.

\bibitem[{Chen et~al.(2022{\natexlab{a}})Chen, Zhang, Rosenberg, Ramabhadran, Moreno, Bapna, and Zen}]{Chen2022MAESTRO}
Zhehuai Chen, Yu~Zhang, Andrew Rosenberg, Bhuvana Ramabhadran, Pedro Moreno, Ankur Bapna, and Heiga Zen. 2022{\natexlab{a}}.
\newblock Maestro: Matched speech text representations through modality matching.
\newblock \emph{arXiv preprint arXiv:2204.03409}.

\bibitem[{Chen et~al.(2022{\natexlab{b}})Chen, Zhang, Rosenberg, Ramabhadran, Moreno, and Wang}]{Chen2022ICASSP}
Zhehuai Chen, Yu~Zhang, Andrew Rosenberg, Bhuvana Ramabhadran, Pedro Moreno, and Gary Wang. 2022{\natexlab{b}}.
\newblock \href {https://doi.org/10.1109/ICASSP43922.2022.9746475} {Tts4pretrain 2.0: Advancing the use of text and speech in asr pretraining with consistency and contrastive losses}.
\newblock In \emph{ICASSP 2022 - 2022 IEEE International Conference on Acoustics, Speech and Signal Processing (ICASSP)}, pages 7677--7681.

\bibitem[{Chung et~al.(2018)Chung, Weng, Tong, and Glass}]{Chung2018NIPS}
Yu-An Chung, Wei-Hung Weng, Schrasing Tong, and James Glass. 2018.
\newblock Unsupervised cross-modal alignment of speech and text embedding spaces.
\newblock \emph{Advances in neural information processing systems}, 31.

\bibitem[{Derczynski et~al.(2017)Derczynski, Nichols, van Erp, and Limsopatham}]{Derczynski2017NUGT}
Leon Derczynski, Eric Nichols, Marieke van Erp, and Nut Limsopatham. 2017.
\newblock Results of the {WNUT2017} shared task on novel and emerging entity recognition.
\newblock In \emph{Proceedings of the 3rd Workshop on Noisy User-generated Text}, pages 140--147.

\bibitem[{Devlin et~al.(2018)Devlin, Chang, Lee, and Toutanova}]{devlin2018bert}
Jacob Devlin, Ming-Wei Chang, Kenton Lee, and Kristina Toutanova. 2018.
\newblock Bert: Pre-training of deep bidirectional transformers for language understanding.
\newblock \emph{arXiv preprint arXiv:1810.04805}.

\bibitem[{Goodfellow et~al.(2020)Goodfellow, Pouget-Abadie, Mirza, Xu, Warde-Farley, Ozair, Courville, and Bengio}]{Goodfellow2020ACM}
Ian Goodfellow, Jean Pouget-Abadie, Mehdi Mirza, Bing Xu, David Warde-Farley, Sherjil Ozair, Aaron Courville, and Yoshua Bengio. 2020.
\newblock Generative adversarial networks.
\newblock \emph{Communications of the ACM}, 63(11):139--144.

\bibitem[{Graves(2012)}]{grave2012}
Alex Graves. 2012.
\newblock \href {http://arxiv.org/abs/1211.3711} {Sequence transduction with recurrent neural networks}.
\newblock {arXiv preprint arXiv:1211.3711}.

\bibitem[{Han et~al.(2020)Han, Zhang, Zhang, Yu, Chiu, Qin, Gulati, Pang, and Wu}]{han20_interspeech}
Wei Han, Zhengdong Zhang, Yu~Zhang, Jiahui Yu, Chung-Cheng Chiu, James Qin, Anmol Gulati, Ruoming Pang, and Yonghui Wu. 2020.
\newblock {ContextNet}: Improving convolutional neural networks for automatic speech recognition with global context.
\newblock In \emph{Interspeech 2020}, pages 3610--3614.

\bibitem[{Ho et~al.(2020)Ho, Jain, and Abbeel}]{ho2020denoising}
Jonathan Ho, Ajay Jain, and Pieter Abbeel. 2020.
\newblock Denoising diffusion probabilistic models.
\newblock In \emph{Advances in Neural Information Processing Systems}, volume~33, pages 6840--6851.

\bibitem[{Ho and Salimans(2022)}]{ho2022classifier}
Jonathan Ho and Tim Salimans. 2022.
\newblock \href {https://arxiv.org/abs/2207.12598} {Classifier-free diffusion guidance}.
\newblock {arXiv preprint arXiv:2207.12598}.

\bibitem[{Hsu et~al.(2021)Hsu, Bolte, Tsai, Lakhotia, Salakhutdinov, and Mohamed}]{Hsu2021ITASL}
Wei-Ning Hsu, Benjamin Bolte, Yao-Hung~Hubert Tsai, Kushal Lakhotia, Ruslan Salakhutdinov, and Abdelrahman Mohamed. 2021.
\newblock {HuBERT}: Self-supervised speech representation learning by masked prediction of hidden units.
\newblock \emph{IEEE/ACM Transactions on Audio, Speech, and Language Processing}, 29:3451--3460.

\bibitem[{Huang et~al.(2020{\natexlab{a}})Huang, Hu, Yeung, and Chen}]{huang2020interspeech}
Wenyong Huang, Wenchao Hu, Yu~Ting Yeung, and Xiao Chen. 2020{\natexlab{a}}.
\newblock {Conv-Transformer Transducer}: Low latency, low frame rate, streamable end-to-end speech recognition.
\newblock In \emph{Interspeech 2020}, pages 5001--5005.

\bibitem[{Huang et~al.(2022)Huang, Zhang, Yeung, Jiang, and Liu}]{huang2022spiral}
Wenyong Huang, Zhenhe Zhang, Yu~Ting Yeung, Xin Jiang, and Qun Liu. 2022.
\newblock \href {https://openreview.net/forum?id=TBpg4PnXhYH} {{SPIRAL}: Self-supervised perturbation-invariant representation learning for speech pre-training}.
\newblock In \emph{International Conference on Learning Representations}.

\bibitem[{Huang et~al.(2020{\natexlab{b}})Huang, Kuo, Thomas, Kons, Audhkhasi, Kingsbury, Hoory, and Picheny}]{Huang2020ICASSP}
Yinghui Huang, Hong-Kwang Kuo, Samuel Thomas, Zvi Kons, Kartik Audhkhasi, Brian Kingsbury, Ron Hoory, and Michael Picheny. 2020{\natexlab{b}}.
\newblock Leveraging unpaired text data for training end-to-end speech-to-intent systems.
\newblock In \emph{2020 IEEE International Conference on Acoustics, Speech and Signal Processing (ICASSP 2020)}, pages 7984--7988.

\bibitem[{Karras et~al.(2022)Karras, Aittala, Aila, and Laine}]{karras2022elucidating}
Tero Karras, Miika Aittala, Timo Aila, and Samuli Laine. 2022.
\newblock \href {https://arxiv.org/abs/2206.00364} {Elucidating the design space of diffusion-based generative models}.
\newblock {arXiv preprint arXiv:2206.00364}.

\bibitem[{Kharitonov et~al.(2023)Kharitonov, Vincent, Borsos, Marinier, Girgin, Pietquin, Sharifi, Tagliasacchi, and Zeghidour}]{kharitonov2023speak}
Eugene Kharitonov, Damien Vincent, Zal{\'a}n Borsos, Rapha{\"e}l Marinier, Sertan Girgin, Olivier Pietquin, Matt Sharifi, Marco Tagliasacchi, and Neil Zeghidour. 2023.
\newblock Speak, read and prompt: High-fidelity text-to-speech with minimal supervision.
\newblock \emph{arXiv preprint arXiv:2302.03540}.

\bibitem[{Kim et~al.(2021)Kim, Kim, Lee, and Ha}]{Kim2021ICASSP}
Minjeong Kim, Gyuwan Kim, Sang-Woo Lee, and Jung-Woo Ha. 2021.
\newblock {ST-BERT}: Cross-modal language model pre-training for end-to-end spoken language understanding.
\newblock In \emph{2021 IEEE International Conference on Acoustics, Speech and Signal Processing (ICASSP 2021)}, pages 7478--7482.

\bibitem[{Laptev et~al.(2020)Laptev, Korostik, Svischev, Andrusenko, Medennikov, and Rybin}]{laptev2020you}
Aleksandr Laptev, Roman Korostik, Aleksey Svischev, Andrei Andrusenko, Ivan Medennikov, and Sergey Rybin. 2020.
\newblock \href {https://arxiv.org/abs/2005.07157} {You do not need more data: Improving end-to-end speech recognition by text-to-speech data augmentation}.
\newblock {arXiv preprint arXiv:2005.07157}.

\bibitem[{Larson and Leach(2022)}]{Larson2022REDWOOD}
Stefan Larson and Kevin Leach. 2022.
\newblock \href {https://arxiv.org/abs/2204.05483} {Redwood: Using collision detection to grow a large-scale intent classification dataset}.
\newblock {arXiv preprint arXiv:2204.05483}.

\bibitem[{Larson et~al.(2019)Larson, Mahendran, Peper, Clarke, Lee, Hill, Kummerfeld, Leach, Laurenzano, Tang et~al.}]{Larson2019Arxiv}
Stefan Larson, Anish Mahendran, Joseph~J Peper, Christopher Clarke, Andrew Lee, Parker Hill, Jonathan~K Kummerfeld, Kevin Leach, Michael~A Laurenzano, Lingjia Tang, et~al. 2019.
\newblock An evaluation dataset for intent classification and out-of-scope prediction.
\newblock \emph{arXiv preprint arXiv:1909.02027}.

\bibitem[{Lewis et~al.(2019)Lewis, Liu, Goyal, Ghazvininejad, Mohamed, Levy, Stoyanov, and Zettlemoyer}]{lewis2019bart}
Mike Lewis, Yinhan Liu, Naman Goyal, Marjan Ghazvininejad, Abdelrahman Mohamed, Omer Levy, Ves Stoyanov, and Luke Zettlemoyer. 2019.
\newblock Bart: Denoising sequence-to-sequence pre-training for natural language generation, translation, and comprehension.
\newblock \emph{arXiv preprint arXiv:1910.13461}.

\bibitem[{Liu et~al.(2020)Liu, Xu, Yu, Dai, Ji, Cahyawijaya, Madotto, and Fung}]{Liu2020CROSSNER}
Zihan Liu, Yan Xu, Tiezheng Yu, Wenliang Dai, Ziwei Ji, Samuel Cahyawijaya, Andrea Madotto, and Pascale Fung. 2020.
\newblock {CrossNER}: Evaluating cross-domain named entity recognition.
\newblock ArXiv preprint arXiv:2012.04373.

\bibitem[{Luan et~al.(2018)Luan, He, Ostendorf, and Hajishirzi}]{Luan2018EMNLP}
Yi~Luan, Luheng He, Mari Ostendorf, and Hannaneh Hajishirzi. 2018.
\newblock Multi-task identification of entities, relations, and coreference for scientific knowledge graph construction.
\newblock In \emph{Proceedings of the 2018 Conference on Empirical Methods in Natural Language Processing}, pages 3219--3232.

\bibitem[{Lugosch et~al.(2020)Lugosch, Meyer, Nowrouzezahrai, and Ravanelli}]{LUGOSH2020ICASSP}
Loren Lugosch, Brett~H Meyer, Derek Nowrouzezahrai, and Mirco Ravanelli. 2020.
\newblock Using speech synthesis to train end-to-end spoken language understanding models.
\newblock In \emph{2020 IEEE International Conference on Acoustics, Speech and Signal Processing (ICASSP 2020)}, pages 8499--8503.

\bibitem[{Lüscher et~al.(2019)Lüscher, Beck, Irie, Kitza, Michel, Zeyer, Schlüter, and Ney}]{L_scher_2019}
Christoph Lüscher, Eugen Beck, Kazuki Irie, Markus Kitza, Wilfried Michel, Albert Zeyer, Ralf Schlüter, and Hermann Ney. 2019.
\newblock {RWTH} asr systems for {LibriSpeech: Hybrid vs Attention}.
\newblock In \emph{Interspeech 2019}, pages 231--235.

\bibitem[{Nichol et~al.(2021)Nichol, Dhariwal, Ramesh, Shyam, Mishkin, McGrew, Sutskever, and Chen}]{nichol2021glide}
Alex Nichol, Prafulla Dhariwal, Aditya Ramesh, Pranav Shyam, Pamela Mishkin, Bob McGrew, Ilya Sutskever, and Mark Chen. 2021.
\newblock \href {https://arxiv.org/abs/2112.10741} {Glide: Towards photorealistic image generation and editing with text-guided diffusion models}.
\newblock {arXiv preprint arXiv:2112.10741}.

\bibitem[{Panayotov et~al.(2015)Panayotov, Chen, Povey, and Khudanpur}]{panayotov2015librispeech}
Vassil Panayotov, Guoguo Chen, Daniel Povey, and Sanjeev Khudanpur. 2015.
\newblock Librispeech: An {ASR} corpus based on public domain audio books.
\newblock In \emph{2015 IEEE international conference on acoustics, speech and signal processing (ICASSP)}, pages 5206--5210. IEEE.

\bibitem[{{Park} et~al.(2020){Park}, {Zhang}, {Chiu}, {Chen}, {Li}, {Chan}, {Le}, and {Wu}}]{park2019specaugment}
D.~S. {Park}, Y.~{Zhang}, C.~{Chiu}, Y.~{Chen}, B.~{Li}, W.~{Chan}, Q.~V. {Le}, and Y.~{Wu}. 2020.
\newblock {SpecAugment} on large scale datasets.
\newblock In \emph{2020 IEEE International Conference on Acoustics, Speech and Signal Processing (ICASSP 2020)}, pages 6879--6883.

\bibitem[{Park et~al.(2020)Park, Zhang, Jia, Han, Chiu, Li, Wu, and Le}]{park2020improved}
Daniel~S. Park, Yu~Zhang, Ye~Jia, Wei Han, Chung-Cheng Chiu, Bo~Li, Yonghui Wu, and Quoc~V. Le. 2020.
\newblock Improved noisy student training for automatic speech recognition.
\newblock In \emph{Interspeech 2020}, pages 2817--2821.

\bibitem[{Peddinti et~al.(2018)Peddinti, Wang, Povey, and Khudanpur}]{Peddinti2018}
Vijayaditya Peddinti, Yiming Wang, Daniel Povey, and Sanjeev Khudanpur. 2018.
\newblock Low latency acoustic modeling using temporal convolution and {LSTMs}.
\newblock \emph{IEEE Signal Processing Letters}, 25(3):373--377.

\bibitem[{Peng et~al.(2020)Peng, Zhu, Li, Li, Li, Zeng, and Gao}]{Peng2020SCGPT}
Baolin Peng, Chenguang Zhu, Chunyuan Li, Xiujun Li, Jinchao Li, Michael Zeng, and Jianfeng Gao. 2020.
\newblock \href {https://doi.org/10.18653/v1/2020.findings-emnlp.17} {Few-shot natural language generation for task-oriented dialog}.
\newblock In \emph{Findings of the Association for Computational Linguistics: EMNLP 2020}, pages 172--182, Online. Association for Computational Linguistics.

\bibitem[{Popov et~al.(2021)Popov, Vovk, Gogoryan, Sadekova, and Kudinov}]{popov2021grad}
Vadim Popov, Ivan Vovk, Vladimir Gogoryan, Tasnima Sadekova, and Mikhail Kudinov. 2021.
\newblock {Grad-TTS}: A diffusion probabilistic model for text-to-speech.
\newblock In \emph{International Conference on Machine Learning (ICML)}, pages 8599--8608.

\bibitem[{Qian et~al.(2021)Qian, Bianv, Shi, Kanda, Shen, Xiao, and Zeng}]{Qian2021ICASSP}
Yao Qian, Ximo Bianv, Yu~Shi, Naoyuki Kanda, Leo Shen, Zhen Xiao, and Michael Zeng. 2021.
\newblock Speech-language pre-training for end-to-end spoken language understanding.
\newblock In \emph{2021 IEEE International Conference on Acoustics, Speech and Signal Processing (ICASSP 2021)}, pages 7458--7462.

\bibitem[{Raffel et~al.(2020)Raffel, Shazeer, Roberts, Lee, Narang, Matena, Zhou, Li, and Liu}]{Raffel2020JMLR}
Colin Raffel, Noam Shazeer, Adam Roberts, Katherine Lee, Sharan Narang, Michael Matena, Yanqi Zhou, Wei Li, and Peter~J. Liu. 2020.
\newblock \href {http://jmlr.org/papers/v21/20-074.html} {Exploring the limits of transfer learning with a unified text-to-text transformer}.
\newblock \emph{Journal of Machine Learning Research}, 21(140):1--67.

\bibitem[{Rastogi et~al.(2019)Rastogi, Zang, Sunkara, Gupta, and Khaitan}]{Rastogi2019ARXIV}
Abhinav Rastogi, Xiaoxue Zang, Srinivas Sunkara, Raghav Gupta, and Pranav Khaitan. 2019.
\newblock \href {https://arxiv.org/abs/1909.05855} {Towards scalable multi-domain conversational agents: The schema-guided dialogue dataset}.
\newblock {arXiv preprint arXiv:1909.05855}.

\bibitem[{Rastogi et~al.(2020)Rastogi, Zang, Sunkara, Gupta, and Khaitan}]{Rastogi2020DSTC8}
Abhinav Rastogi, Xiaoxue Zang, Srinivas Sunkara, Raghav Gupta, and Pranav Khaitan. 2020.
\newblock \href {https://arxiv.org/abs/2002.01359} {Schema-guided dialogue state tracking task at dstc8}.
\newblock {arXiv preprint arXiv:2002.01359}.

\bibitem[{Rombach et~al.(2022)Rombach, Blattmann, Lorenz, Esser, and Ommer}]{rombach2022high}
Robin Rombach, Andreas Blattmann, Dominik Lorenz, Patrick Esser, and Bj{\"o}rn Ommer. 2022.
\newblock High-resolution image synthesis with latent diffusion models.
\newblock In \emph{Proceedings of the IEEE/CVF Conference on Computer Vision and Pattern Recognition}, pages 10684--10695.

\bibitem[{Ronneberger et~al.(2015)Ronneberger, Fischer, and Brox}]{ronneberger2015u}
Olaf Ronneberger, Philipp Fischer, and Thomas Brox. 2015.
\newblock U-net: Convolutional networks for biomedical image segmentation.
\newblock In \emph{International Conference on Medical image computing and computer-assisted intervention}, pages 234--241.

\bibitem[{Saharia et~al.(2022)Saharia, Chan, Saxena, Li, Whang, Denton, Ghasemipour, Ayan, Mahdavi, Lopes et~al.}]{saharia2022photorealistic}
Chitwan Saharia, William Chan, Saurabh Saxena, Lala Li, Jay Whang, Emily Denton, Seyed Kamyar~Seyed Ghasemipour, Burcu~Karagol Ayan, S~Sara Mahdavi, Rapha~Gontijo Lopes, et~al. 2022.
\newblock \href {https://arxiv.org/abs/2205.11487} {Photorealistic text-to-image diffusion models with deep language understanding}.
\newblock {arXiv preprint arXiv:2205.11487}.

\bibitem[{Sato et~al.(2022)Sato, Komori, Mishima, Kawai, Mochizuki, Sato, and Ogawa}]{sato2022text}
Hiroaki Sato, Tomoyasu Komori, Takeshi Mishima, Yoshihiko Kawai, Takahiro Mochizuki, Shoei Sato, and Tetsuji Ogawa. 2022.
\newblock Text-only domain adaptation based on intermediate ctc.
\newblock In \emph{Proceedings of the Annual Conference of the International Speech Communication Association, INTERSPEECH}, volume 2022, pages 2208--2212.

\bibitem[{Sharma et~al.(2021)Sharma, Madhavi, and Li}]{Sharma2021ICASSP}
Bidisha Sharma, Maulik Madhavi, and Haizhou Li. 2021.
\newblock Leveraging acoustic and linguistic embeddings from pretrained speech and language models for intent classification.
\newblock In \emph{2021 IEEE International Conference on Acoustics, Speech and Signal Processing (ICASSP 2021)}, pages 7498--7502.

\bibitem[{Shen et~al.(2021)Shen, Hsu, Ray, and Jin}]{Shen2021ACL}
Yilin Shen, Yen-Chang Hsu, Avik Ray, and Hongxia Jin. 2021.
\newblock Enhancing the generalization for intent classification and out-of-domain detection in slu.
\newblock In \emph{Proceedings of the 59th Annual Meeting of the Association for Computational Linguistics and the 11th International Joint Conference on Natural Language Processing (Volume 1: Long Papers)}, pages 2443--2453.

\bibitem[{Song et~al.(2020)Song, Sohl-Dickstein, Kingma, Kumar, Ermon, and Poole}]{song2020score}
Yang Song, Jascha Sohl-Dickstein, Diederik~P Kingma, Abhishek Kumar, Stefano Ermon, and Ben Poole. 2020.
\newblock Score-based generative modeling through stochastic differential equations.
\newblock \emph{arXiv preprint arXiv:2011.13456}.

\bibitem[{Sowa{\'n}ski and Janicki(2020)}]{Sowanski2020ICTSD}
Marcin Sowa{\'n}ski and Artur Janicki. 2020.
\newblock Leyzer: A dataset for multilingual virtual assistants.
\newblock In \emph{International Conference on Text, Speech, and Dialogue}, pages 477--486.

\bibitem[{Sun et~al.(2020)Sun, Zhang, Weiss, Cao, Zen, Rosenberg, Ramabhadran, and Wu}]{sun2020generating}
Guangzhi Sun, Yu~Zhang, Ron~J Weiss, Yuan Cao, Heiga Zen, Andrew Rosenberg, Bhuvana Ramabhadran, and Yonghui Wu. 2020.
\newblock Generating diverse and natural text-to-speech samples using a quantized fine-grained {VAE} and autoregressive prosody prior.
\newblock In \emph{2020 IEEE International Conference on Acoustics, Speech and Signal Processing (ICASSP 2020)}, pages 6699--6703.

\bibitem[{Thomas et~al.(2022)Thomas, Kingsbury, Saon, and Kuo}]{thomas2022integrating}
Samuel Thomas, Brian Kingsbury, George Saon, and Hong-Kwang~J Kuo. 2022.
\newblock Integrating text inputs for training and adapting rnn transducer asr models.
\newblock In \emph{ICASSP 2022-2022 IEEE International Conference on Acoustics, Speech and Signal Processing (ICASSP)}, pages 8127--8131. IEEE.

\bibitem[{Thomas et~al.(2021)Thomas, Kuo, Saon, T{\"u}ske, Kingsbury, Kurata, Kons, and Hoory}]{Thomas2021ICASSP}
Samuel Thomas, Hong-Kwang~J Kuo, George Saon, Zolt{\'a}n T{\"u}ske, Brian Kingsbury, Gakuto Kurata, Zvi Kons, and Ron Hoory. 2021.
\newblock Rnn transducer models for spoken language understanding.
\newblock In \emph{ICASSP 2021-2021 IEEE International Conference on Acoustics, Speech and Signal Processing (ICASSP)}, pages 7493--7497. IEEE.

\bibitem[{Tian and Gorinski(2020)}]{Tian2020IESICR}
Yusheng Tian and Philip~John Gorinski. 2020.
\newblock Improving end-to-end speech-to-intent classification with {Reptile}.
\newblock In \emph{Interspeech 2020}, pages 891--895.

\bibitem[{Tian et~al.(2019)Tian, Yi, Tao, Bai, and Wen}]{tian2019self}
Zhengkun Tian, Jiangyan Yi, Jianhua Tao, Ye~Bai, and Zhengqi Wen. 2019.
\newblock Self-attention transducers for end-to-end speech recognition.
\newblock In \emph{Interspeech 2019}, pages 4395--4399.

\bibitem[{Tjong Kim~Sang and De~Meulder(2003)}]{Sang2003CONLL2003}
Erik~F. Tjong Kim~Sang and Fien De~Meulder. 2003.
\newblock \href {https://aclanthology.org/W03-0419} {Introduction to the {C}o{NLL}-2003 shared task: Language-independent named entity recognition}.
\newblock In \emph{Proceedings of the Seventh Conference on Natural Language Learning at {HLT}-{NAACL} 2003}, pages 142--147.

\bibitem[{Tomasello et~al.(2022)Tomasello, Shrivastava, Lazar, Hsu, Le, Sagar, Elkahky, Copet, Hsu, Adi, Algayres, Nguyen, Dupoux, Zettlemoyer, and Mohamed}]{Tomasello2022STOP}
Paden Tomasello, Akshat Shrivastava, Daniel Lazar, Po-Chun Hsu, Duc Le, Adithya Sagar, Ali Elkahky, Jade Copet, Wei-Ning Hsu, Yossi Adi, Robin Algayres, Tu~Ahn Nguyen, Emmanuel Dupoux, Luke Zettlemoyer, and Abdelrahman Mohamed. 2022.
\newblock \href {https://arxiv.org/abs/2207.10643} {{STOP}: A dataset for spoken task oriented semantic parsing}.
\newblock {arXiv preprint arXiv:2207.10643}.

\bibitem[{Tur et~al.(2010)Tur, Hakkani-T{\"u}r, and Heck}]{Tur2010SLTW}
Gokhan Tur, Dilek Hakkani-T{\"u}r, and Larry Heck. 2010.
\newblock What is left to be understood in {ATIS}?
\newblock In \emph{2010 IEEE Spoken Language Technology Workshop (SLT)}, pages 19--24.

\bibitem[{Vaswani et~al.(2017)Vaswani, Shazeer, Parmar, Uszkoreit, Jones, Gomez, Kaiser, and Polosukhin}]{Vaswani2017}
Ashish Vaswani, Noam Shazeer, Niki Parmar, Jakob Uszkoreit, Llion Jones, Aidan~N Gomez, \L~ukasz Kaiser, and Illia Polosukhin. 2017.
\newblock \href {https://proceedings.neurips.cc/paper/2017/file/3f5ee243547dee91fbd053c1c4a845aa-Paper.pdf} {Attention is all you need}.
\newblock In \emph{Advances in Neural Information Processing Systems}, volume~30.

\bibitem[{Wang et~al.(2020)Wang, Wei, Cao, Xie, and Nie}]{WANG2020ICASSP}
Pengwei Wang, Liangchen Wei, Yong Cao, Jinghui Xie, and Zaiqing Nie. 2020.
\newblock Large-scale unsupervised pre-training for end-to-end spoken language understanding.
\newblock In \emph{2020 IEEE International Conference on Acoustics, Speech and Signal Processing (ICASSP 2020)}, pages 7999--8003.

\bibitem[{Wang et~al.(2021)Wang, Boumadane, and Heba}]{Wang2021HUBERTPARTIAL}
Yingzhi Wang, Abdelmoumene Boumadane, and Abdelwahab Heba. 2021.
\newblock \href {https://arxiv.org/abs/2111.02735} {A fine-tuned wav2vec 2.0/hubert benchmark for speech emotion recognition, speaker verification and spoken language understanding}.
\newblock {arXiv preprint arXiv:2111.02735}.

\bibitem[{Watanabe et~al.(2022)Watanabe, Hori, Karita, Hayashi, Nishitoba, Unno, {Enrique Yalta Soplin}, Heymann, Wiesner, Chen, Renduchintala, and Ochiai}]{Watanabe2018EspnetLibrispeech100h}
Shinji Watanabe, Takaaki Hori, Shigeki Karita, Tomoki Hayashi, Jiro Nishitoba, Yuya Unno, Nelson {Enrique Yalta Soplin}, Jahn Heymann, Matthew Wiesner, Nanxin Chen, Adithya Renduchintala, and Tsubasa Ochiai.
\newblock \href {https://github.com/espnet/espnet/blob/master/egs/librispeech\_100/asr1/RESULTS.md} {{Official Results of Conformer-based Models from ESPNET}} [online]. 2022.

\bibitem[{Weischedel et~al.(2013)Weischedel, Palmer, Marcus, Hovy, Pradhan, Ramshaw, Xue, Taylor, Kaufman, Franchini et~al.}]{Weischedel2013LDC}
Ralph Weischedel, Martha Palmer, Mitchell Marcus, Eduard Hovy, Sameer Pradhan, Lance Ramshaw, Nianwen Xue, Ann Taylor, Jeff Kaufman, Michelle Franchini, et~al. 2013.
\newblock Ontonotes release 5.0 {LDC2013T19}.
\newblock Linguistic Data Consortium.

\bibitem[{Yeh et~al.(2019)Yeh, Mahadeokar, Kalgaonkar, Wang, Le, Jain, Schubert, Fuegen, and Seltzer}]{yeh2019transformer}
Ching-Feng Yeh, Jay Mahadeokar, Kaustubh Kalgaonkar, Yongqiang Wang, Duc Le, Mahaveer Jain, Kjell Schubert, Christian Fuegen, and Michael~L Seltzer. 2019.
\newblock \href {https://arxiv.org/abs/1910.12977} {{Transformer-Transducer}: End-to-end speech recognition with self-attention}.
\newblock {arXiv preprint arXiv:1910.12977}.

\bibitem[{Zang et~al.(2020)Zang, Rastogi, Sunkara, Gupta, Zhang, and Chen}]{Zang2020ACL}
Xiaoxue Zang, Abhinav Rastogi, Srinivas Sunkara, Raghav Gupta, Jianguo Zhang, and Jindong Chen. 2020.
\newblock {MultiWOZ 2.2}: A dialogue dataset with additional annotation corrections and state tracking baselines.
\newblock In \emph{Proceedings of the 2nd Workshop on Natural Language Processing for Conversational AI, ACL 2020}, pages 109--117.

\bibitem[{Zhang et~al.(2023)Zhang, Li, Zhang, Zhan, Wang, Zhou, and Qiu}]{zhang2023speechgpt}
Dong Zhang, Shimin Li, Xin Zhang, Jun Zhan, Pengyu Wang, Yaqian Zhou, and Xipeng Qiu. 2023.
\newblock Speechgpt: Empowering large language models with intrinsic cross-modal conversational abilities.
\newblock \emph{arXiv preprint arXiv:2305.11000}.

\bibitem[{{Zhang} et~al.(2020){Zhang}, {Lu}, {Sak}, {Tripathi}, {McDermott}, {Koo}, and {Kumar}}]{zhang2020transformer}
Q.~{Zhang}, H.~{Lu}, H.~{Sak}, A.~{Tripathi}, E.~{McDermott}, S.~{Koo}, and S.~{Kumar}. 2020.
\newblock {Transformer Transducer}: A streamable speech recognition model with transformer encoders and {RNN-T} loss.
\newblock In \emph{2020 IEEE International Conference on Acoustics, Speech and Signal Processing (ICASSP 2020)}, pages 7829--7833.

\bibitem[{Zhang et~al.(2022{\natexlab{a}})Zhang, Chen, Zhou, Wu, Ren, Liu, Yao, Gong, Dai, Li et~al.}]{Zhang2022SPEECHLM}
Ziqiang Zhang, Sanyuan Chen, Long Zhou, Yu~Wu, Shuo Ren, Shujie Liu, Zhuoyuan Yao, Xun Gong, Lirong Dai, Jinyu Li, et~al. 2022{\natexlab{a}}.
\newblock \href {https://arxiv.org/abs/2209.15329} {{SpeechLM}: Enhanced speech pre-training with unpaired textual data}.
\newblock {arXiv preprint arXiv:2209.15329}.

\bibitem[{Zhang et~al.(2022{\natexlab{b}})Zhang, Zhou, Ao, Liu, Dai, Li, and Wei}]{Zhang2022EMNLP}
Ziqiang Zhang, Long Zhou, Junyi Ao, Shujie Liu, Lirong Dai, Jinyu Li, and Furu Wei. 2022{\natexlab{b}}.
\newblock \href {https://aclanthology.org/2022.emnlp-main.108} {{S}peech{UT}: Bridging speech and text with hidden-unit for encoder-decoder based speech-text pre-training}.
\newblock In \emph{Proceedings of the 2022 Conference on Empirical Methods in Natural Language Processing}, pages 1663--1676, Abu Dhabi, United Arab Emirates. Association for Computational Linguistics.

\end{thebibliography}
\bibliographystyle{acl_natbib}

\appendix

\end{document}